\documentclass{article}

\usepackage{PRIMEarxiv}

\usepackage[utf8]{inputenc} 
\usepackage[T1]{fontenc}    
\usepackage{hyperref}       
\usepackage{url}            
\usepackage{booktabs}       
\usepackage{amsfonts}       
\usepackage{nicefrac}       
\usepackage{microtype}      
\usepackage{lipsum}
\usepackage{fancyhdr}       
\usepackage{graphicx}       
\graphicspath{{media/}}     

\usepackage[numbers]{natbib}

\usepackage{multirow}
\usepackage{float}
\usepackage{subfig} 
\usepackage{amsmath}
\usepackage{mathtools}
\usepackage{bm}
\usepackage{multirow}
\usepackage{enumitem}
\usepackage[ruled,vlined,linesnumbered,noend]{algorithm2e}
\usepackage[table,xcdraw]{xcolor}

\usepackage{combelow}

\usepackage{xcolor}

\definecolor{MidnightBlue}{rgb}{0.1, 0.1, 0.44}

\title{
Efficient Machine Learning Ensemble Methods for Detecting Gravitational Wave Glitches in LIGO Time Series
}

\author{
  Elena-Simona Apostol and Ciprian-Octavian Truic{\u{a}}\\
  National University of Science and Technology Politehnica Bucharest, Bucharest, Romania \\
  \texttt{
    elena.apostol@upb.ro, 
    ciprian.truica@upb.ro
    }
}

\begin{document}
\maketitle

\begin{abstract}
The phenomenon of Gravitational Wave (GW) analysis has grown in popularity as technology has advanced and the process of observing gravitational waves has become more precise. 
Although the sensitivity and the frequency of observation of GW signals are constantly improving, the possibility of noise in the collected GW data remains.
In this paper, we propose two new Machine and Deep learning ensemble approaches (i.e., ShallowWaves and DeepWaves Ensembles) for detecting different types of noise and patterns in datasets from GW observatories.
Our research also investigates various Machine and Deep Learning techniques for multi-class classification and provides a comprehensive benchmark, emphasizing the best results in terms of three commonly used performance metrics (i.e., accuracy, precision, and recall). 
We train and test our models on a dataset consisting of annotated time series from real-world data collected by the Advanced Laser Interferometer GW Observatory (LIGO).
We empirically show that the best overall accuracy is obtained by the proposed DeepWaves Ensemble, followed close by the ShallowWaves Ensemble.
\end{abstract}


\maketitle

\section{Introduction}
The Advanced Laser Interferometer Gravitational Wave Observatory (LIGO)~\cite{Aasi2015Ligo} and the Advanced Virgo~\cite{Acernese2014Virgo} conducted the first observations in 2015 to prove the existence of Gravitational Waves (GW) from a binary black hole (BBH) coalescence.
Terrestrial interferometers, such as LIGO or Virgo, search for signatures of gravitational radiation that are usually mixed with lots of background noise. 
The hardest to remove are the noises that mimic real astrophysical signals and/or influence the quality of the received signal data, referred to herein as glitches.

There are many processes in the universe that produce detectable gravitational waves. 
The current Earth-based GW detectors search in the frequency range from 10 Hz to 10 kHz for gravitational signals. 
The main issue with the received signals is that the data may contain a high number of glitches. 
In this context, we consider a glitch as being a transient noise event that mimics real astrophysical signals and/or influences the quality of data. 
As a result, many detection algorithms were considered in order to optimize the LIGO/Virgo data cleansing process. 
Most of the current detection algorithms used for GW data cleansing belong to the matched-filtering category which momentarily targets only a subset of the dimensional space available from the GW detector. 
These algorithms are also extremely sensitive, which means that they can be subject to errors. 
Therefore, new algorithms must be investigated and further expanded and optimized in order to target other types of searches. 
Some of the most recent solutions apply Machine Learning or Deep Learning algorithms for efficiently handling complex GW datasets.

In this article, we attempt to solve the problem of multi-class classification of different glitches from signals collected from GW generators.
Although there are some Machine/Deep Learning (ML/DL) solutions for GW detection, the majority of these solutions address the problem as a simple binary classification, e.g., signal or noise.
We also discovered several articles (e.g.,~\cite{Sakai2022, Fernandes2023}) that use multi-class classification for glitches detection, but they only evaluate classical ML algorithms or architectures based on Convolutional Neural Networks (CNN) or basic Artificial Neural Networks (ANN), without testing or addressing other types of networks, such as recurrent neural networks.

\paragraph*{Problem Definition}
We consider a stream of multi-variant time series data $t_i \in \mathbb{R}^n$ collected from a GW generator.
Each record from the time series, i.e., $t_i$, can be an astrophysical signal, an astrophysical signal plus noise, or noise alone. 
Each reading has a set of features $t_i = \{\Sigma p_i\}$, where 
$p_i$ is a feature of the gravitational wave, e.g., the frequency, bandwidth, SNR (signal-to-noise ratio), and duration. 

Let $C = \{c_i\}$ be a family of classes for different glitches (e.g., Power Line glitches) and Scattered Light noise induced by seismic activity. 
Each glitch family is mainly identified by its unique morphology of time-frequency~\cite{ligoOnline2023}.

The goal is to design multi-class classification models $\delta: \mathbb{R}^n \rightarrow C$ for anomaly detection 
that (1) can be efficiently implemented and distributed, and (2) have good statistical performance even when using small to medium unbalanced datasets.

\paragraph*{Objectives} The main objective of this article is to apply different supervised Machine Learning and Deep Learning algorithms for anomaly detection in order to detect glitches and new classes of glitches in the LIGO time series data.
In this paper, we propose two new ensemble models (ShallowWaves and DeepWaves Ensembles) that have good accuracy in classifying different types of glitches in GW. 
We also present an extensive benchmark for the multi-class classification problem on 15 Machine Learning and 9 Deep Learning models that detect noise in the LIGO dataset.
The experiments are done using Stratified k-Folds to split the dataset into training and testing sets.
We use grid search to extract the best hyperparameter for the classical Machine Learning algorithms. 
Although the obtained results have good accuracy for the classical ML algorithms, we observed that the performance of the neural networks differs depending on the type of Deep Network architecture applied.
The best overall accuracy is obtained by the DeepWaves Ensemble, followed closely by the ShallowWaves Ensemble.

In the last decade, new HPC applications have emerged that require both streaming and high computing capabilities. 
Detecting anomalies, i.e., glitches, in the field of Gravitational Wave astronomy is such an HPC solution that requires the online processing of large datasets collected from Earth-based GW detectors.
For such applications, we need a Big data stream analysis solution.
In our architecture, we choose MPI and MPI Streaming library in order to efficiently distribute our problem. 

To summarize our contributions:
\begin{itemize}
    \item [\textit{(1)}] We propose two new anomaly detection ensemble models, i.e., ShallowWaves Ensemble and DeepWaves Ensemble.
    \item [\textit{(2)}] We perform an in-depth analysis of our models and compare our results with traditional Machine Learning and Deep Learning algorithms.
    \item [\textit{(3)}] We analyze the use of different Deep Learning models, i.e., Recurrent Neural Networks and Deep Belief Networks, which have not been tried before for detecting anomalies (i.e., glitches) in GW.
    \item [\textit{(4)}]We efficiently distribute our solution using MPI and MPI Streaming.
\end{itemize}

This work was done as part of G2net Cost Action~\cite{G2NetCOST} that aims at creating a broad network of scientists from four different areas of expertise, i.e., GW physics, Geophysics, Computing Science, and Robotics, with the shared goal of addressing challenges in data analysis and noise characterization for GW detectors.

The rest of this paper is structured as follows.
Section~\ref{sec:rw} presents a survey of several state-of-the-art Machine Learning methods for analyzing Gravitational Waves datasets.
In Section~\ref{sec:metodology}, we discuss the baseline and the basic models used in our proposed solution.
Section~\ref{sec:ourSolution} presents the architecture of our distributed system and discusses the proposed Machine and Deep Learning models.
Section~\ref{sec:results} showcases and discusses the obtained results.
Lastly, in Section~\ref{sec:conclusions} we conclude and we present several new directions and improvements for the proposed solution.

\section{Related Work}~\label{sec:rw}

Recent research on developing novel algorithms for optimizing the search for gravitational waves and signal glitches has led to the integration of machine learning (ML) techniques into the field of astrophysics. 
To emphasize the effectiveness of such a solution, this section focuses on analyzing related articles that use Machine Learning and Deep Learning algorithms in the field of Gravitational Waves. 

There are quite a few Machine and Deep Learning based solutions for gravitational wave analysis. 
Some solutions focus on gravitational wave detection and parameter identification~\cite{George2018, Yan2022}, while others try to identify glitches~\cite{Corizzo2020,Szczepanczyk2023}.

George et al.~\cite{George2018} pioneered the use of Deep Learning with Convolutional Neural Networks (CNN) that take time series as inputs for the fast detection and modeling of gravitational wave signals. 
They developed an extension to the existing approach called Deep Filtering.
Their results show that Deep Filtering achieves similar sensitivities and lower errors compared to matched-filtering while being far more computationally efficient and more resilient to glitches, allowing real-time processing of weak time series signals in non-stationary and non-Gaussian noise with minimal resources. 

Gabbard et al.~\cite{Gabbard2018} use a single deep CNN to detect binary black hole Gravitational Wave signals. 
They demonstrate that the CNN approach closely matches the sensitivity of matched-filtering for all datasets across the range of false alarm probabilities explored in the analysis.

Yan et al.~\cite{Yan2022} try to identify gravitational waves using multilayer perceptron neural network architectures with ReLU (Rectified Linear Unit) activation function. 
Their network structure is obtained basically by replacing the max module in matched filtering with a multilayer perceptron network.

Sasaoka et al.~\cite{Sasaoka2022} use Temporal Convolutional Networks (TCN) and Artificial Neural Networks (ANN) for sky localization of gravitational waves.
Each GW signal is classified into one of 18 sectors. 
For training and testing, they use a labeled synthetic dataset generated by the GGWD tool~\cite{ggwd2019}. 

Mitra et al.~\cite{Mitra2023} generate GW signals for a range of stellar models using numerical simulations and apply machine learning, i.e., Random Forest, to train and classify the signals. 

Corizzo et al.~\cite{Corizzo2020} propose two approaches (i.e., supervised and unsupervised) involving deep auto-encoder models to analyze time series collected from GW detectors.
The proposed approaches are used for the binary classification of the data as noise or real signals.
As noted by the authors, they achieved high scalability using the Apache Spark framework.

Szczepa{\'{n}}czyk et al.~\cite{Szczepanczyk2023} present a solution that searches for generic short-duration gravitational-wave (GW) transients (or GW bursts) in the data from the third observing run of Advanced LIGO and Advanced Virgo. 
They use the WaveBurst pipeline enhanced with a decision-tree classification algorithm for more efficient separation of GW signals from noise.

Although most ML-based solutions on GW are only applied as binary classification problems (e.g., detect noise or signal), there are a few articles that tackle more complex multi-class classification problems, i.e., detect different classes of signals or noises.
Sakai et al.~\cite{Sakai2022} propose an unsupervised learning architecture for classifying the noise of interferometric Gravitational Wave detectors. 
The proposed architecture uses a deep CNN to classify use a Deep CNN architecture to classify transient noises from Gravitational Spy dataset\footnote{\url{https://www.zooniverse.org/projects/zooniverse/gravity-spy}} that classify the noises in 22 classes. 
Fernandes et al.~\cite{Fernandes2023} also investigate the use of CNN (including the ConvNeXt network family) to classify different types of glitches and gravitational waves in time series data from the Advanced LIGO detectors. 
Mesuga et al.\cite{Mesuga2021} also use the Gravitational Spy dataset that contains images and apply different CNN-based classifiers, e.g., VGG-19 a 19 layers deep CNN, ResNet-101 an 101 layers deep CNN.

To our knowledge, the majority of the current solutions use either pure synthetic data (e.g., ~\cite{Gabbard2018, Sasaoka2022}) or a combination of real and synthetic data (e.g., ~\cite{Yan2022, Corizzo2020, George2018}). 
The major problem with using synthetic data is the high possibility of creating bias or overfitting the model. 
This problem is usually not even discussed in the corresponding papers.
For example, Corizzo et al.~\cite{Corizzo2020} do not discuss the potential overfitting of the model introduced by the choice of replicating multiple times all negative class time series available from the original dataset.

\section{Methodology}~\label{sec:metodology}
This section aims to create an overall picture of the models used and the reasons behind the proposed pipelines from Section~\ref{sec:ourSolution}.
In our experiments, we also use these models as baselines.

\textbf{K-Nearest Neighbor (KNN)} is a lazy learning method in the sense that no model is learned from the training data, learning only occurs when a test example needs to be classified. 
KNN works as follows: \textit{i)} use the distance function to compute the distance between each point in the dataset, \textit{ii)} select nearest points (called the k-nearest neighbors), \textit{iii)} assign to each test point the most frequent class from the set of its k nearest neighbor.

\textbf{Gaussian Naïve Bayes (GNB)} classifier is a probabilistic classification algorithm that computes the probability of each element of the dataset belonging to a class using Bayes theorem.
When dealing with continuous data, a typical assumption is that the continuous values associated with each class are distributed according to a Gaussian distribution~\cite{Strickland2016}. 

\textbf{Logistic regression (LogReg)} estimates the probability of an independent variable to fall into a certain level of the categorical response given a set of predictors.

\textbf{Decision Tree} Classifier builds a classification model in the form of a tree structure. 
It breaks down iteratively a dataset into smaller and smaller subsets, using a top-down approach, while at the same time, an associated decision tree is incrementally developed. 
At each iteration, the splitting is done by selecting the attribute for which the smaller datasets (i.e., partitions) obtain the highest homogeneity, i.e., contain instances with similar values.
There are multiple approaches for computing the homogeneity of a partition.
In this work, we use Entropy and Information Gain, utilizing the \textit{C4.5} algorithm, as well as the Gini Impurity, employing the \textit{CART} algorithm.

The final result is a tree with decision nodes and leaf nodes. 
The decision nodes are attributes.
Branches refer to discrete values (one or more) or intervals for these attributes. 
The leaves are labeled with classes.

\textbf{Random Forest (RF)} is an ensemble classifier that grows multiple trees at the same time to solve the problem of structural similarity given by the greedy approach of decision trees.
Thus, Random Forest learns multiple decision tree models for which the resulting predictions are less correlated.
To achieve this, the algorithm changes the feature space at the moment of splitting.
Instead of having access to all the characteristics, the decision tree has access to a limited number of features selected at random.

\textbf{Extremely Randomized Trees (ERT)} add another step of randomization in the way splits are computed.
It essentially consists of randomizing strongly both attribute and cut-point choice while splitting a tree node. 
In the extreme case, it builds totally randomized trees whose structures are independent of the output values of the learning sample. 
The strength of the randomization can be tuned to problem specifics by the appropriate choice of a parameter.

\textbf{Adaptive Boosting (AdaBoost)} uses weak learners’ output and combines it into a weighted sum that represents the final output of the classifier. 
It constructs a strong classifier as a linear combination of weak classifiers. 
AdaBoost assigns a ``weight" to each training example, which determines the probability that each example should appear in the training set. 
After training a classifier, AdaBoost increases the weight on the misclassified examples so that these examples will make up a larger part of the next classifier’s training set, and hopefully, the next classifier trained will perform better on them.

\textbf{Gradient Boosted (XGBoost)} is a technique similar to AdaBoost that creates multiple weak classifiers to improve accuracy.
The technique uses the gradient descent method to minimize the prediction error of each weak classifier based on the previous ones.
In this work, we use CART as the weak classifier with 3 types of boosters: gbtree, dart, and gblinear.
The bgtree booster minimizes the loss function using a gradient descent algorithm, dart is a similar version to bgtree that uses dropout techniques to avoid overfitting, and gblinear uses generalized linear regression instead of decision trees.
Using these 3 different boosters, we train 3 Gradient Boosted Trees using the CART algorithm: XGBoost-gbtree, XGBoost-dart, XGBoost-linear.

\textbf{Perceptron} is a simple Neural Network (NN) that contains only one layer, which is the output layer. 
It processes the input data, encoded in a way that facilitates the computation, using the equations presented at logistic regression. 
The function used is called the activation function and it is selected based on the result that should be achieved. The most popular ones are the sigmoid function, the hyperbolic tangent (tanh), the rectified linear unit (ReLU), and the leaky ReLU. 
The first two are suitable for binary classification, whereas the last two are used when the output needs to be limited to positive values.

\textbf{Multi-Layer Perceptron (MLP)} is a NN algorithm that stacks multiple layers of perceptrons in a fully connected network to solve a classification problem. 
The layers are divided into an input layer, multiple hidden layers, and an output layer. 	
The input layer and the output layer are required, but the hidden layers are optional. 
Usually, the number of hidden layers is small and, depending on the problem, an optimal number can be found for achieving the best accuracy of the network. 
Each layer can have a different number of neurons and computes the result based on the input from the previous layer, with the exception of the input layer.

\textbf{Long Short-Term Memory network (LSTM)}~\cite{Hochreiter1997} is an extension of the Recurrent Neural Network (RNN) that uses two state components for classification.
The first component is a short-term memory that learns the short-term dependency between the previous state and the current state.
The second component is a long-term memory, representing a cell's internal state that stores long-term dependency between the previous and current state.
LSTM uses three gates to preserve the long-term memory in the state:
(1) input gate ($i \in \mathbb{R}^{n}$),
(2) forget gate ($f \in \mathbb{R}^{n}$), and
(3) output gate ($o \in \mathbb{R}^{n}$).

In this paper, we chose to test and incorporate this type of recurrent network in our solution because LSTM manages to avoid some of the issues of other recurrent networks, i.e., the vanishing and the exploding gradient. 
It does this by regulating the recurrent weights learning process.

\textbf{Convolutional Neural Network (CNN)} is a class of NN mainly used in visual imagery and is inspired by the connectivity pattern of neurons in the visual cortex. 
Individual neurons respond to stimuli only in a restricted region of the visual field and a collection of said fields overlap to cover the entire visual area.
A CNN consists of an input layer and an output layer, as well as multiple hidden layers: a convolutional layer, a pooling layer, and a fully connected layer. 
The fully connected layer is exactly like the hidden layer in regular neural networks.
The convolution layer of the neural network is the core building block in CNNs that does most of the computation. 
The parameters for this layer consist of a set of learnable filters, spatially small along width and height, but which extend through the whole depth of the input (in text classification, the depth is 1). 
This filter can be seen as a window sliding across the input, and the only part of the input that is analyzed at a certain time is the one inside the window. 
The result of the analysis is, actually, the dot product between the entries in the filter and the input at the current position of the window.

\textbf{Deep Belief Networks (DBN)}~\cite{Hua2015deep} are a class of feedforward deep neural networks that are composed of multiple hidden layers with connections between the layers but not between units within each layer. 
DBNs are a sophisticated type of generative neural network that uses an unsupervised machine learning model to produce results. 

\textbf{Ensemble methods} combine multiple classifiers to obtain a better one. 
Combined classifiers are similar (use the same learning method) but the training datasets or the weights’ examples are different.
Ensemble methods use Bagging (Bootstrap Aggregation) and Boosting.
Bagging~\cite{Breiman1996} is a machine learning ensemble algorithm designed to improve the stability and accuracy of machine learning that constructs a training set from the initial labeled data set by sampling with replacement. Bagging reduces the variance and helps to avoid overfitting.
Boosting~\cite{Schapire1990} consists of building a sequence of weak classifiers and adding them to the structure of the final strong classifier.

\section{Proposed Solution}~\label{sec:ourSolution}

In this section, we describe our distributed Master-Worker architecture and proposed novel ensemble models used to detect different types of anomalies in data from Earth-based Gravitational Wave (GW) detectors.

\begin{figure}[!htbp]
    \centering
    \includegraphics[width=1\columnwidth]{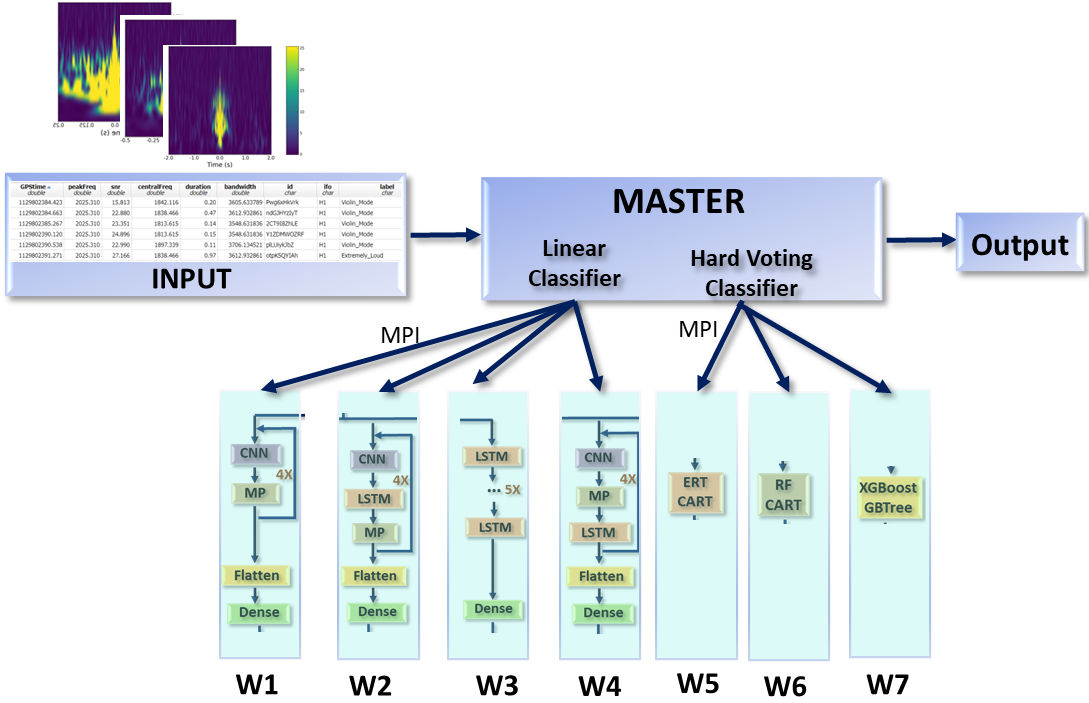}
    \caption{Proposed Architecture}
    \label{fig:archi}
\end{figure}

Figure~\ref{fig:archi} depicts our proposed distributed architecture for online processing GW time series.
The architecture consists of a master node and several worker nodes (i.e., $W_i$).
A time series reading is received by the Master and sent using an MPI Streaming API to all the Workers in the cluster. 
Each Worker has a saved Machine Learning model, representing one of the branches from our proposed ensemble models, i.e., ShallowWaves Ensemble and DeepWaves Ensemble.

\begin{figure}[!htbp]
    \centering
    \includegraphics[width=0.7\columnwidth]{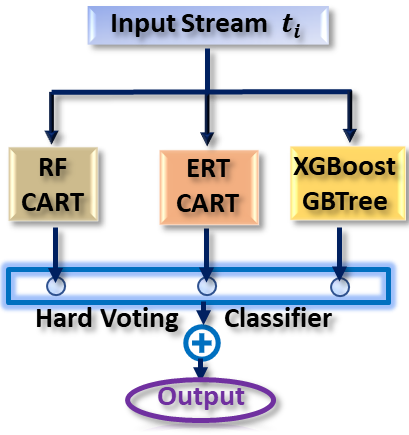}
    \caption{ShallowWaves Ensemble Model}
    \label{fig:ShalowWaves}
\end{figure}

\paragraph*{ShallowWaves Ensemble}  consists of three Machine Learning branches and a Hard Voting classifier, as depicted in Figure~\ref{fig:ShalowWaves}. 
The algorithms employed are RF CART, ERT CART, and XGBoost-gbtree. 
Each algorithm grows CART trees using either bagging or gradient boosting techniques.
The predictor collects the results of each model and, by employing a hard voting approach, makes the final decision.
By taking this approach, ShallowWaves Ensemble improves anomaly detection accuracy by minimizing False Positive and False Negative predictions.
For our implementation, we use the classical classifiers from the  \href{https://scikit-learn.org}{Sklearn} library and the XGBClassifier from the \href{https://xgboost.readthedocs.io/}{xgboost} library.

\begin{figure}[!htbp]
    \centering
    \includegraphics[width=0.9\columnwidth]{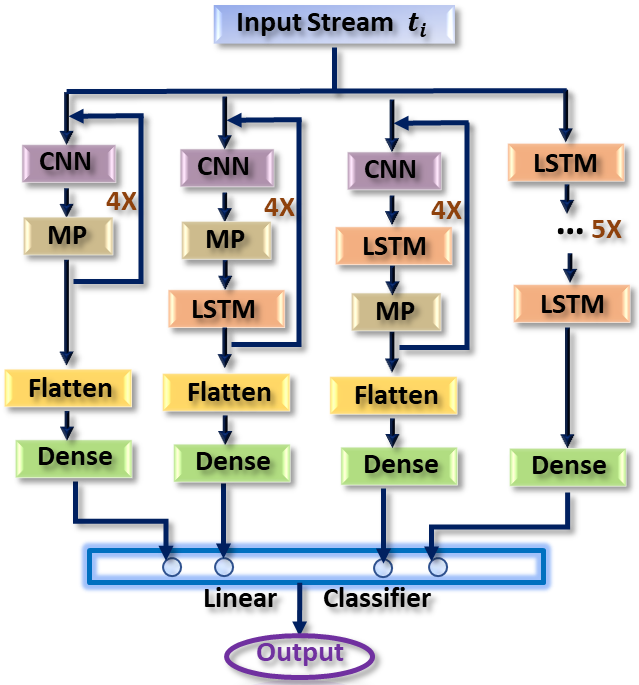}
    \caption{DeepWaves Ensemble Model}
    \label{fig:DeepWaves}
\end{figure}

\paragraph*{DeepWaves Ensemble} consists of four Deep Learning branches (i.e., $4 \times (CNN+MP)$, $4 \times (CNN+MP+LSTM)$, $4\times(CNN+LSTM+MP)$,  $5\times LSTM$) and a Linear classifier (Figure~\ref{fig:DeepWaves}).
This ensemble combines CNN and LSTM layers to improve the performance of the classifiers obtained when using each branch individually.
The first branch employs CNN and MaxPolling layers to extract and create informative representations of the data.
The next two branches use different configurations of CNN and LSTM layers to extract new features and improve the architecture's overall accuracy as they are used in cases when the dataset size is small.
The last branch uses stacked LSTM layers to create a hierarchical feature representation of the input data.
The results of these layers are combined using a Linear layer which gives the final prediction.
We use the \href{https://keras.io/}{Keras} library to implement our models and the \href{https://scikit-learn.org}{Sklearn} library for preprocessing and metric calculations.

\section{Experimental Results}~\label{sec:results}
In this section, we present the experimental setup and the used dataset, and we analyze the obtained results for the anomaly detection task.
The source code for our models is publicly available on GitHub at \url{https://github.com/DS4AI-UPB/ML4GW/}.

\subsection{Experimental Setup}
The experiments are done on a cluster with 6 nodes currently running Ubuntu 22.04 x64, each with 1 Intel Core i7-4790S CPU with 8 cores at 3.20GHz, 16 GB RAM, and 500GB HDD. 
On each node, we have OpenMPI v4.0.6\footnote{Open MPI~\url{https://www.open-mpi.org/}} and the corresponding host configuration file.
Our code source is written in Python version 3.11.
MPI for Python package\footnote{MPI for Python package~\url{https://mpi4py.readthedocs.io/en/stable}} was used to provide Python bindings for the MPI.
We use the MPI Stream library~\cite{Peng2017mpi} to support data streams between the Master and the Workers.

\subsection{Dataset}

This research has made use of data obtained from the Gravitational Wave Open Science Center~\cite{AllDatasets2023}.
The dataset is from the first observing run (O1)~\cite{LigoO1Dataset2016} of the Advanced LIGO, i.e., Hanford (H), and Livingston (L) detectors, from September 2015 to December 2015.

\begin{table}[!htbp]
\centering
\caption{Glitch Classes for the Anomaly Detection}
\label{tab:glitches-classes}
\resizebox{\columnwidth}{!}{
\begin{tabular}{lll}
\hline
\textbf{Label} & \textbf{\#} & \textbf{Description} \\ \hline
    Scattered Light         &     427        &             low-frequency, long duration, humpy         \\
    Power Line      &     450        &            narrow in frequency, last for $\approx 0.2-0.5$ s          \\
    1080Lines   &      4       &          steady stream, around 1080 Hz            \\
    1400Ripples	&        83     &        short-duration, around 1400 Hz           \\
    Air Compressor	&      57       &         short-duration glitches - around 50 Hz         \\
    Blip	&        1763     &           teardrop' shape, 30 - 500 HZ           \\
    Repeating Blips	&    91         &       blip glitches that repeat               \\
    Violin Mode	&      137       &     short and dot-like                 \\
    Whistle	&     146        &          W or V shape            \\
    Scratchy	&     269        &       short-duration repeating, intermediate freq.               \\
    Helix	&       270      &        resemble a vortex, intermediate freq.           \\
    Light Modulation	&      400       &     several bright spikes in close succession                 \\
    Low Frequency Burst	&      527       &          loud, short-lived, low-freq.             \\
    Wandering Line	&    21         &         lines, long duration, meander in freq.             \\
    Koi Fish	&     709        &          similar to Blip, resemble a fish in shape          \\
    Low Frequency Lines	&       494      &      horizontal lines at low freq.                \\
    Chirp	&    60         &         sweeping upwards in freq. over time             \\
    Extremely Loud	&     448        &        from a major disturbance in the detectors              \\
    Paired Doves	&    26         &    repeating glitches, alternate increasing-decreasing freq.                  \\
    Tomte	&       93      &       lower-freq., usually triangular in shape               \\
    No Glitch	&       41      &         no apparent transient noise structure            \\
    None of the Above	&       151      &   a catch-all for glitches that do not fit             \\ \hline
\end{tabular}
}
\end{table}

\begin{figure}[!htbp]
    \centering
    \includegraphics[width=1\columnwidth]{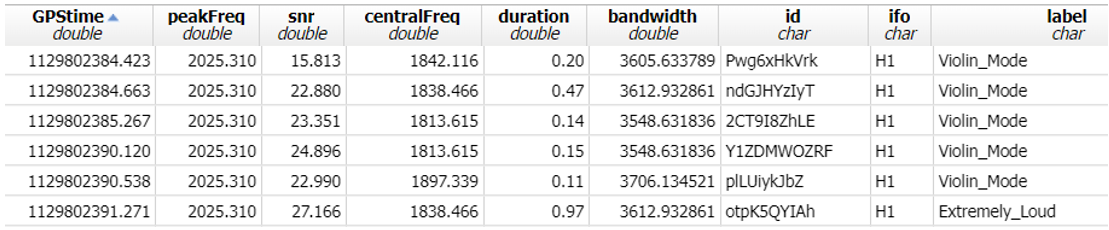}
    \caption{Sample from the dataset}
    \label{fig:sampledata}
\end{figure}

\begin{table*}[!htbp]
\centering
\caption{Classical Machine Learning models - Anomaly detection results}
\label{tab:MLresults}
\begin{tabular}{lccccc}
\hline
\textbf{Algorithm} & \textbf{Accuracy} & \textbf{\begin{tabular}[c]{@{}l@{}}Weighted Precision\end{tabular}} & \textbf{\begin{tabular}[c]{@{}l@{}}Micro Precision\end{tabular}} & \textbf{\begin{tabular}[c]{@{}l@{}}Weighted Recall\end{tabular}} & \textbf{\begin{tabular}[c]{@{}l@{}}Micro Recall\end{tabular}} \\ \hline
KNN  & $0.70\pm0.01$ & $0.70\pm0.01$ & $0.70 \pm 0.01$  & $0.70\pm0.01$  & $0.70\pm0.01$   \\
GNB & $0.63 \pm 0.01$ & $0.66 \pm 0.01$ & $0.63\pm 0.01$ & $0.63\pm 0.01$ & $0.63 \pm 0.01$ \\
LogReg & $0.61\pm0.01$ & $0.52\pm0.02$ & $0.61\pm0.01$ & $0.61\pm0.01$ & $0.61\pm0.01$ \\
CART & $0.78\pm0.01$ & $0.78\pm0.01$ & $0.78\pm0.01$ & $0.78\pm0.01$ & $0.78\pm0.01$ \\
C4.5 & $0.79\pm0.01$ & $0.79\pm0.01$ & $0.79\pm0.01$ & $0.79\pm0.01$ & $0.79\pm0.01$ \\
AdaBoost CART & $0.78\pm0.01$  & $0.79\pm0.01$ & $0.78\pm0.01$ & $0.78\pm0.01$ & $0.78\pm0.01$ \\
AdaBoost C4.5 & $0.79\pm0.01$ & $0.79\pm0.01$ & $0.79\pm0.01$ & $0.79\pm0.01$ & $0.79\pm0.01$ \\
RF CART & $0.84\pm0.01$ & $0.83\pm0.01$ & $0.84\pm0.01$ & $0.84\pm0.01$ & $0.84\pm0.01$ \\ 
RF C4.5 & $0.84\pm0.01$ & $0.84\pm0.01$ & $0.84\pm0.01$ & $0.84\pm0.01$ & $0.84\pm0.01$ \\
ERT CART & $0.84\pm0.01$ & $0.83\pm0.01$ & $0.84\pm0.01$ & $0.84\pm0.01$ & $0.84\pm0.01$ \\
ERT C4.5 & $0.84\pm0.01$ & $0.83\pm0.01$ & $0.84\pm0.01$ & $0.84\pm0.01$ & $0.84\pm0.01$ \\
XGBoost-gbtrees & $0.84\pm0.01$ & $0.84\pm0.01$ & $0.84\pm0.01$ & $0.84\pm0.01$ & $0.84\pm0.01$ \\
XGBoost-gblinear  & $0.61\pm0.01$ & $0.53\pm0.02$ & $0.61\pm0.01$ & $0.61\pm0.01$ & $0.61\pm0.01$ \\
XGBoost-dart & $0.84\pm0.01$ & $0.84\pm0.01$ & $0.84\pm0.01$ & $0.84\pm0.01$ & $0.84\pm0.01$ \\
\textcolor{MidnightBlue}{\textbf{ShallowWaves Ensemble}} & $0.89\pm0.02$ & $0.88\pm0.03$ & $0.89\pm0.02$ & $0.89\pm0.02$ & $0.89\pm0.02$ \\ 
\hline
\end{tabular}
\end{table*}

\begin{table*}[!htbp]
\centering
\caption{Deep Learning models - Anomaly detection results}
\label{tab:DLresults}
\begin{tabular}{lccccc}
\hline
\textbf{Algorithm} & \textbf{Accuracy} & \textbf{\begin{tabular}[c]{@{}l@{}}Weighted Precision\end{tabular}} & \textbf{\begin{tabular}[c]{@{}l@{}}Micro Precision\end{tabular}} & \textbf{\begin{tabular}[c]{@{}l@{}}Weighted Recall\end{tabular}} & \textbf{\begin{tabular}[c]{@{}l@{}}Micro Recall\end{tabular}} \\ \hline
Perceptron & $0.74\pm0.09$ & $0.77\pm0.04$ & $0.74\pm0.09$ & $0.74\pm0.09$ & $0.74\pm0.09$ \\
MLP & $0.74\pm0.03$ & $0.63\pm0.07$  & $0.74\pm0.03$ & $0.74\pm0.03$ & $0.74\pm0.03$ \\ 
DBN & $0.60\pm0.01$ & $0.53\pm0.02$ & $0.60\pm0.01$  & $0.60\pm0.01$ & $0.60\pm0.01$ \\
LSTM & $0.81\pm0.03$ & $0.74\pm0.07$ & $0.81\pm0.03$  & $0.81\pm0.03$ & $0.81\pm0.03$ \\
CNN  & $0.71\pm0.05$ & $0.63\pm0.07$ & $0.71\pm0.05$ & $0.71\pm0.05$ & $0.71\pm0.05$ \\
5 $\times$ LSTM & $0.83\pm0.01$ & $0.73\pm0.02$ & $0.83\pm0.01$ & $0.83\pm0.01$ & $0.83\pm0.01$ \\
4 $\times$ (CNN + MP) & $0.81\pm0.02$ & $0.75\pm0.03$ & $0.81\pm0.02$ & $0.81\pm0.02$ & $0.81\pm0.02$ \\
4 $\times$ (CNN + LSTM + MP) & $0.84\pm0.03$ & $0.80\pm0.06$ & $0.84\pm0.03$ & $0.84\pm0.03$ & $0.84\pm0.03$ \\
4 $\times$ (CNN + MP + LSTM) & $0.86\pm0.03$ & $0.81\pm0.06$ & $0.86\pm0.03$ & $0.86\pm0.03$ & $0.86\pm0.03$ \\
\textcolor{MidnightBlue}{\textbf{DeepWaves Ensemble}} & $\bm{0.91\pm0.03}$ & $0.87\pm0.05$ & $0.91\pm0.03$ & $0.91\pm0.03$ & $0.91\pm0.03$ \\ 
\hline
\end{tabular}
\end{table*}

\begin{table*}[!htbp]
\centering
\caption{Best classification models - Anomaly detection results}
\label{tab:results_best}
\begin{tabular}{lccccc}
\hline
\textbf{Algorithm} & \textbf{Accuracy} & \textbf{\begin{tabular}[c]{@{}l@{}}Weighted Precision\end{tabular}} & \textbf{\begin{tabular}[c]{@{}l@{}}Micro Precision\end{tabular}} & \textbf{\begin{tabular}[c]{@{}l@{}}Weighted Recall\end{tabular}} & \textbf{\begin{tabular}[c]{@{}l@{}}Micro Recall\end{tabular}} \\ \hline
RF CART & $0.84\pm0.01$ & $0.83\pm0.01$ & $0.84\pm0.01$ & $0.84\pm0.01$ & $0.84\pm0.01$ \\ 
RF C4.5 & $0.84\pm0.01$ & $0.84\pm0.01$ & $0.84\pm0.01$ & $0.84\pm0.01$ & $0.84\pm0.01$ \\
ERT CART & $0.84\pm0.01$ & $0.83\pm0.01$ & $0.84\pm0.01$ & $0.84\pm0.01$ & $0.84\pm0.01$ \\
ERT C4.5 & $0.84\pm0.01$ & $0.83\pm0.01$ & $0.84\pm0.01$ & $0.84\pm0.01$ & $0.84\pm0.01$ \\
XGBoost-gbtree & $0.84\pm0.01$ & $0.84\pm0.01$ & $0.84\pm0.01$ & $0.84\pm0.01$ & $0.84\pm0.01$ \\
XGBoost-dart & $0.84\pm0.01$ & $0.84\pm0.01$ & $0.84\pm0.01$ & $0.84\pm0.01$ & $0.84\pm0.01$ \\
4 $\times$ (CNN + LSTM + MP) & $0.84\pm0.03$ & $0.80\pm0.06$ & $0.84\pm0.03$ & $0.84\pm0.03$ & $0.84\pm0.03$ \\
4 $\times$ (CNN + MP + LSTM) & $0.86\pm0.03$ & $0.81\pm0.06$ & $0.86\pm0.03$ & $0.86\pm0.03$ & $0.86\pm0.03$ \\
\textcolor{MidnightBlue}{\textbf{ShallowWaves Ensemble}} & $0.89\pm0.02$ & $0.88\pm0.03$ & $0.89\pm0.02$ & $0.89\pm0.02$ & $0.89\pm0.02$ \\ 
\textcolor{MidnightBlue}{\textbf{DeepWaves Ensemble}} & $\bm{0.91\pm0.03}$ & $0.87\pm0.05$ & $0.91\pm0.03$ & $0.91\pm0.03$ & $0.91\pm0.03$ \\ 
\hline
\end{tabular}
\end{table*}

The O1 dataset consists of 6667 entries and is annotated using 22 classes of different transient noise with non-stationary and non-Gaussian features.
The annotation process was done by the Gravity Spy project\footnote{\url{https://www.zooniverse.org/projects/zooniverse/gravity-spy}}, a citizen science project hosted by the Zooniverse platform. 
In Table~\ref{tab:glitches-classes}, we present each class by providing a short description and the number of contained items. 
As can be seen, the dataset is highly unbalanced, e.g., we have 1763 Blip entries and only 4 1080Lines entries.

The dataset has a total of 8 features and 1 label which we use for training and predicting. 
The features specific to a dataset entry are the time at which the entry was detected (\textit{GPStime}), the peak frequency of the gravitational wave spectrum (\textit{peakFreq}), the signal-to-noise ratio (\textit{snr}),  the central frequency of the wavelet (\textit{centralFreq}), \textit{duration}, \textit{bandwidth}, \textit{id}, and the interferometer which captured the entry (\textit{ifo}).
A sample of the dataset is provided in Figure~\ref{fig:sampledata}.

\subsection{Anomaly Detection Results}

To compare our proposed ensemble solutions, we train 14 Machine Learning models and 9 Deep Learning models to act as a baseline. 
To fine-tune the hyperparameters of the classical Machine Learning algorithms, we employ grid search.  
As the number of parameters increases exponentially for the Deep Neural Networks with each layer, we could not fine-tune the hyperparameters. 
For all our experiments, we use Stratified k-Folds to split the data into the testing and training sets with $k=10$.

We have computed for each experiment the Accuracy, Weighted Precision, Micro Precision, Weighted Recall, and Micro Recall. 
The evaluation of the methods also considers the class weights. 

Table~\ref{tab:MLresults} presents the results obtained using the Machine Learning algorithms.
The decision trees algorithms, i.e., CART and C4.5, as well as the baseline ensemble methods that use multiple decision trees, i.e., AdaBoost, Random Forests (RF), Extremely Randomized Trees (ERT), and XGBoost-gbtree, present high accuracy. 
Furthermore, XGBoost-gbree and XGBoost-dart have the same performance, while XGBoost-gblinear has a lower performance.
The proposed classical Machine Learning ensemble, i.e., ShallowWaves, obtains the overall best accuracy when compared with the baseline.

To determine how DeepWaves Ensemble performs, we first analyze the results obtained by each branch individually using ablation testing (Table~\ref{tab:DLresults}.
We also compare the results obtained by the proposed DeepWaves Ensemble with the results obtained when training other deep learning methods, i.e., Perceptron, MLP, DBN, LSTM, and CNN.
The Multi-Layer Perceptron (MLP) uses 5 layers with a different number of units, alternating the activation function between ReLU and softmax.
We see no significant difference between the results obtained by a single Perceptron and the MLP.
The Deep Belief Network (DBN) obtains the overall worst results. 
This shows that for our highly unbalanced medium-sized dataset, this complex architecture does not manage to find good parameters.

Between the Deep Neural Networks baseline, the best performance is obtained by 4 $\times$ (CNN + MP + LSTM).
The performance slightly decreases when we stack multiple LSTM layers (5 $\times$ LSTM), but it shows some performance improvement over the LSTM model. 

For the CNN architecture, we also used a MaxPooling Layer (MP). 
The worst performance between the models tested is obtained when using a single CNN layer followed by an MP layer.
The performance increases when we add more layers, i.e., 4 x (CNN + MP).
By adding LSTM layers, i.e., 4 $\times$ (CNN + LSTM + MP) and 4 $\times$ (CNN + MP + LSTM), the performance increases even more, obtaining better results than the ones obtained with the CNN and 4 $\times$ (CNN + MP) models. 

Overall, the proposed DeepWaves Ensemble manages to outperform all the other models (Table~\ref{tab:results_best}).
By building an ensemble with multiple Deep Neural Network architectures, we manage to train a model that:
(1) minimizes the False Positives and False Negatives, increasing the scores of all the evaluation metrics used;
(2) creates a better representation of the feature space; and
(3) learns hidden features to improve the dissimilarities among classes.

Thus, we can conclude that for such a highly unbalanced dataset, classical ML models perform in some cases better than some DL models that do not manage to find good parameters.
Also, we generally need a data-driven approach to select the best model and the best hyperparameters for a specific dataset. 
In this situation, an ensemble model is the most suitable solution.

\section{Conclusions}~\label{sec:conclusions}
Glitches represent noise events that are similar to real gravitational waves. They have the capability of influencing the actual data by being falsely considered Gravitational Waves (GW). 
In this paper, we analyze 14 Machine Learning models and 9 Deep Learning models that detect glitches in GW data.
We also propose two new models,  ShallowWaves -- a Machine Learning ensemble model -- and  DeepWaves -- a Deep Learning ensemble model.
We distribute our models in a private cluster using MPI.

From our experiments, we conclude that the best overall accuracy is obtained by the proposed DeepWaves Ensemble, followed close by the ShallowWaves Ensemble.
Also, adding a more complex Deep Learning architecture does not necessarily improve accuracy, but it can drastically increase the time performance.

Due to the highly unbalanced and relatively small dataset, some of the neural networks obtain worse performance metrics than some classical Machine Learning algorithms. 
In future work, we plan to have an in-depth analysis of this and consider testing all the models using data from the second (O2) and third observing run (O3) of the Advanced LIGO detector.

\section*{Acknowledgment}
This work was partially funded by the National University of Science and Technology Politehnica Bucharest through PubArt, EU COST Action CA17137 A network for Gravitational Waves, Geophysics, and Machine Learning (G2Net), ``A network for Gravitational Waves, Geophysics and Machine Learning'' project (COST CA17137 STSM 45382) and ``Efficient ML Algorithms for Detecting Glitches and Data Patterns in LIGO Time Series'' project (COST CA17137 STSM 46260).

\bibliographystyle{plainnat}  
\bibliography{main}

\end{document}